# Hypothesis Management in Situation-Specific Network Construction

**Kathryn Blackmond Laskey**
Systems Engineering & Operations
Research Department
George Mason University, MS 4A6
Fairfax, VA  22030
klaskey@gmu.edu

**Suzanne M. Mahoney**
Information Extraction and
Transport, Inc.
1911 N. Ft Myer Dr., Suite 600
Arlington, VA  22209
suzanne@iet.com

**Ed Wright**
Information Extraction and
Transport, Inc.
1911 N. Ft Myer Dr., Suite 600
Arlington, VA  22209
ewright@iet.com

## Abstract

This paper considers the problem of knowledge-based model construction in the presence of uncertainty about the association of domain entities to random variables. *Multi-entity Bayesian networks* (MEBNs) are defined as a representation for knowledge in domains characterized by uncertainty in the number of relevant entities, their interrelationships, and their association with observables. An MEBN implicitly specifies a probability distribution in terms of a hierarchically structured collection of Bayesian network fragments that together encode a joint probability distribution over arbitrarily many interrelated hypotheses. Although a finite query-complete model can always be constructed, association uncertainty typically makes exact model construction and evaluation intractable. The objective of hypothesis management is to balance tractability against accuracy. We describe an approach to hypothesis management, present an application to the problem of military situation awareness, and compare our approach to related work in the tracking and fusion literature.

## 1   INTRODUCTION

Knowledge based model construction (KBMC) is required for complex problems in which it is infeasible to specify *a priori* a Bayesian network encompassing all the situations one might encounter in problem solving. Such problems typically involve an unbounded number of entities of different types interacting with each other in varied ways, giving rise to observable indicators that cannot be unambiguously associated with the domain entities generating them. Patterns of entity structure, behavior and relationships can be encoded as fragments of Bayesian networks and stored in knowledge bases as objects or frames (Laskey and Mahoney, 1997; Koller and Pfeffer, 1997; Bangsø and Wuillemin, 2000). KBMC systems build problem-specific models from such knowledge bases. A KBMC system includes a knowledge base, search operators for retrieving problem-relevant knowledge base elements, network construction operators, network evaluation operators, and model construction control mechanisms.

Objectives for a KBMC system are to minimize costs of representation, retrieval, construction and evaluation, while providing accurate responses to queries. Mahoney and Laskey (1998) defined a situation-specific network as a minimal query-complete network constructed from a knowledge base in response to a query for the probability distribution on a set of target variables given evidence and context variables. In domains for which the situation-specific network may be intractably large, the problem is to approximate the situation-specific network with a high-value network, where value incorporates considerations of time and complexity of construction and evaluation, as well as accuracy of results on the target query.

Hypothesis management is particularly important when the number of entities, their types and their interrelationships cannot be specified *a priori*. We describe Multi-Entity Bayesian Networks (MEBNs), an extension of Bayesian networks to incorporate uncertainty about the number and types of entities and how they are related to each other. We introduce a special symbol *, meaning "not relevant to query," to refer to consequences derived from incorrect hypotheses about which entities are present and how they are related to each other. The use of the * value in hypothesis management is discussed.

We illustrate our approach using a simplified problem from the military situation assessment domain. The objective is to infer the presence and activities of military company sized maneuver units which may be armor companies, mechanized infantry companies, or company teams. Our inferences are based on observations of military vehicles: tanks, armored personnel carriers (APC), and trucks. These observations may be incomplete, inconsistent and inaccurate. A company has 2 or 3 maneuver platoons and a company headquarters (HQ) platoon. Companies may be armor, mechanized infantry, or mixed, depending on the composition of the constituent maneuver and company headquarters (HQ) platoons. Maneuver platoons usually have 4 vehicles, although there may be fewer. Armor platoons are composed of tanks; mechanized infantry platoons are composed of APCs. The armor company HQ platoon has 2 tanks, an APC, and two trucks. The mechanized infantry company HQ platoon has 3 APCs and 3 trucks. Vehicles from the company HQ platoons routinely visit the other platoons of the company. The maneuver units carry out activities, which (greatly simplified) fall into two states: conducting combat operations, or refit / preparation for combat. In addition they may be moving around the battlefield. The formation and activity of the maneuver platoons depends on their activity and the activity of their parent unit.

Our system takes inputs from systems which process raw sensor data (primarily radio frequency, radar, and imagery) into reports indicating hypothesized vehicles and activities, together with probabilistic qualifiers on the hypothesized vehicle types and activities. Our system takes these pre-processed reports as evidence from which to infer groups and their activities. This paper presents a simplified example that illustrates key issues that arise in hypothesis management for situation assessment.

This paper is organized as follows. Section 2 provides background on knowledge based model construction and related work on hypothesis management in multi-target tracking. Section 3 briefly describes MEBNs and presents an illustrative set of fragments from our problem domain. Section 4 modifies the network construction algorithm presented in Mahoney and Laskey (1999) to: (1) permit entities to be hypothesized to fill all undesignated slots in network fragments; and (2) incorporate uncertainty about which entities are relevant to a query and how these entities are related to each other. Section 5 describes how hypothesis management is incorporated into knowledge based model construction. The paper concludes with a summary and discussion section.

## 2. BACKGROUND

### 2.1 Hypothesis management in KBMC systems

A standard Bayesian network defines a joint probability distribution over attributes of entities, where the entities are assumed to be exchangeable individuals in a population of interest. For example, in a medical diagnosis domain, the entities are patients and the random variables refer to attributes such as background information, symptoms, diseases, and test results. This type of model is too restrictive for problems in which neither the number of relevant entities nor the relevant attributes and relationships can be specified in advance. For such problems, several authors (e.g., Laskey and Mahoney, 1997; Mahoney and Laskey, 1998; Pfeffer, et al., 1999; Bangsø and Wuillemin, 2000; Haddawy, 1994; Charniak and Goldman, 1993; Wellman, et al., 1992) have suggested encoding knowledge as a collection of partially specified probability models that can be assembled at run-time into a problem-specific Bayesian network.

Very little attention has been devoted to the problem of uncertainty in the association of domain entities to random variables in the model. The language and construction operators in SPOOK support reference and structural uncertainty, but published work (e.g., Pfeffer et al, 1999; Bangsø and Wuillemin, 2000) does not discuss hypothesis management in any detail. Mahoney and Laskey (1998) note the importance of managing hypotheses in the presence of structural uncertainty, but did not propose a method for treating the hypothesis management problem. Goldman's (1990) plan recognition system included a hypothesis management component. An enhanced marker-passer (Carroll and Charniak 1991) used a small Bayesian network to rapidly identify high probability candidate explanations for patterns of evidence. A more extensive Bayesian network was then constructed to evaluate explanations proposed by the marker-passer. After construction, the network was simplified by declaring highly probable hypotheses as evidence and pruning low-probability portions of the network. Goldman noted that locality assumptions (e.g. temporal and spatial coincidence of evidence) were needed to ensure that high probability explanations would be identified by the marker passer.

### 2.3 Hypothesis Management in Tracking Systems

There is an extensive literature in the tracking community on the problem of tracking multiple targets in the presence of association uncertainty (Stone et al., 1999). In multi-target tracking, observations on a set of targets arrive in an ordered discrete sequence. The objective is to identify the number of targets and their states (e.g., target type, position, velocity). The tracking problem is usually decomposed into data association, state estimation, and hypothesis management.

Data association is the problem of associating observations with hypothesized targets. Given a set of observations and a set of hypothesized entities, data association identifies which hypothesized targets might have given rise to each observation. Typically data association involves application of fast thresholding or "gating" operators, followed by a more expensive and more accurate calculation on associations that meet the threshold. In single-hypothesis systems, each observation is associated with a single, best-fitting track or rejected as a false alarm. In multiple-hypothesis systems, multiple association hypotheses are maintained simultaneously.

State estimation uses the reports associated with a hypothesized target to estimate its position, velocity, target type, and possibly other state variables, and to project the target state forward in time to the next set of reports.

Because the number of association hypotheses grows exponentially with the number of observations, techniques have been developed to keep the number of hypotheses manageable. Hypothesis management initiates new hypothesized targets, prunes existing hypothesized targets, and prunes or combines association hypotheses. Decisions about track initiation, pruning and combination commonly involve applying thresholds to rapidly reject highly improbable hypotheses, and then applying more computationally intensive evaluation methods to hypotheses that pass the threshold.

## 3. MULTI-ENTITY BNs

*Multi-entity Bayesian networks* (MEBNs) represent knowledge for domains in which the relevant entities, attributes and relationships cannot be specified in advance of receiving a query. An MEBN implicitly specifies a probability distribution over a constructively defined, dynamically extensible hypothesis space consisting of variable-length tuples of hierarchically organized entities. MEBNs have first-order expressive power. They represent uncertainty not only about attributes of and relationships among domain entities, but also about which entities, attributes and relationships are relevant to a given query. The probability model for an MEBN is specified as a collection of Bayesian network fragments (Laskey and Mahoney, 1997) that together implicitly encode a joint probability distribution over the sample space. A query to an MEBN consists of a request to compute the joint probability distribution for a set of target random variable instances given the values of a set of evidence random variable instances. Inference in an MEBN can be performed by applying knowledge-based model construction (Wellman, et al., 1992) to construct a situation-specific Bayesian network (Mahoney and Laskey, 1998). For problems involving association uncertainty, the situation-specific network is typically intractable and hypothesis management is required to balance computational load against response accuracy.

Entities are categorized into types.[1] Features and behaviors of entities of the same type are represented in an MEBN as exchangeable random variables. These random variables and probability distributions are composed recursively using fragments of directed graphical models.

*Definition 1:* A *hypothesis type*[2] has a unique type label $T$, a non-empty set of *identifying attributes* $(a_0,\ldots,a_k)$, and a non-empty set $\mho_T$ of atomic values. Each identifying attribute $a_i$ has an attribute name $L_i$ and a hypothesis type $T_i$. There is a special *hypothesis identifier* type $H_{ID}$ with a single identifying attribute $a_0$ called the *identifier* that takes on values in a countable *hypothesis instance identifier set* $\Im$. The attribute $a_0$ in every hypothesis type is called the identifier and is of type $H_{ID}$.

$\Rightarrow$ *Definition 2:* A *hypothesis instance* of type $T$ has an *attribute instance label* for each identifying attribute, and a *value* $v \in \mho_T$. The attribute instance label $l_j$ for identifying attribute $a_j$ is the value of the identifier of a hypothesis instance of type $T_j$, of the type associated with $a_j$. The value $i \in \Im$ taken on by the identifier attribute $a_0$ is shared by no other hypothesis instance.

Hypothesis types can be implemented in software as frame or object classes. Hypothesis instances can be implemented as frame or object instances. In an MEBN, hypothesis instances of a given type are represented as exchangeable random variables, where the identifying attributes of the hypothesis type are the unit of replication. Structured hypothesis types can be built up using Bayesian network fragments (Laskey and Mahoney, 1997).

---

[1] Subtypes and inheritance are naturally treated within the MEBN formalism, but are not discussed in this paper for reasons of clarity and space.

[2] Hypothesis types were called random variable classes in Laskey and Mahoney (1997). Traditionally, random variables are defined as functions on a fixed, *a priori* given, sample space. In an MEBN, the sample space is constructed recursively from the hypothesis value sets. Hypothesis types are then random variables that map elements of the recursively defined sample space to random variable instances.

*Definition 3:* A *BN fragment type* consists of a fragment ID, a set of *fragment identifying attributes*, a set of *input hypothesis types*, a non-empty set of *resident hypothesis types*, a *correspondence function*, a *fragment graph*, and a *local distribution* for each resident hypothesis type. Each identifying attribute has a label and an associated hypothesis type. The fragment graph is an acyclic directed graph containing a node for each input and each resident hypothesis type. Nodes corresponding to input hypothesis types must be roots in the fragment graph. The local distribution for each resident hypothesis type specifies a set of probability distributions over possible values of the resident type. For root nodes there is a single distribution; for non-root nodes there is one distribution for each combination of values of the node's parents. The correspondence function maps each identifying attribute of each resident and input hypothesis type to a fragment identifying attribute of matching type. A type-consistent assignment of hypothesis instances to identifying attributes defines an *instance* of the fragment type.

A BN fragment is a hypothesis type, with identifying attributes equal to the fragment identifying attributes and value set equal to the cross product of the values sets of the resident and input hypothesis types. All instances of a fragment encode identical conditional distributions on their resident hypothesis instances given their input hypothesis instances. This definition of network fragment is less general than the definition given in Laskey and Mahoney (1997), where influence combination methods were used to construct distributions for nodes with parents in different fragments. This extension is non-essential to the purposes of this paper and is therefore omitted for simplicity of exposition.

Figure 1 shows a set of BN fragments for the problem of inferring the activities and unit types (armor, mechanized, or mixed) of platoon sized groups and the companies they belong to. Each node is represented by a text label and one or more identifying attributes in parentheses. Input hypothesis types are shaded gray; resident hypothesis types are unshaded. The dark gray nodes in the upper left corner of some fragments are called *association hypotheses*. Association hypotheses represent uncertainty about which hypothesis instance is assigned to identifying attributes that are referred to by a parent node but not by its child. To understand why it is necessary to represent association hypotheses, consider fragment *F10*, which relates a report on a unit's activity to the actual activity the unit is engaging in. The fragment is conditioned on the assumption that the unit generating the report $r$ is the same unit $u$ whose activity is being described. Uncertainty about which unit generated the report is represented by the node $U(r)$. The dotted line from the conditioning hypothesis $U(r)=u$ to ReportedActivity($r$) means that the local distribution for the ReportedActivity(*r*) node applies only when the unit *u* is equal to the unit *U*(*r*) that generated report *r*.

Network fragments contain association hypotheses for any identifying attribute that appears in a node but not in one of its children. A dotted line is drawn from an association hypothesis to any node not containing its identifying attribute but having a parent that does. The dotted line indicates that the distribution for the node is conditioned on the association hypothesis. Fragments containing association hypotheses are called *association conditioned BN fragments*.

***Definition 4*:** An *association conditioned BN fragment F* is a BN fragment together with a set of zero or more *association hypothesis types*. There is an association hypothesis type *Z* for each identifying attribute *z* for which there exists a resident node *B* not referring to *z* but having a parent that refers to *z*. The association hypothesis *Z* for identifying attribute *z* may not have *z* as one of its identifying attributes. *Z* takes as possible values identifiers of hypothesis instances of the type associated with identifying attribute *z*. The distribution encoded by *F* is conditioned on *Z* taking on the value assigned to identifying attribute *z* when an instance of *F* is created.

***Definition 5*:** A *multi-entity Bayesian network* consists of a finite set $\mathcal{K}$ of hypothesis types and a finite set $\mathcal{F}$ of association conditioned BN fragments defined on these hypothesis types, such that $\mathcal{K}$ and $\mathcal{F}$ satisfy the following conditions:

1. Each hypothesis type is resident in no more than one BN fragment.

2. Each hypothesis type appearing in some fragment is resident in at least one fragment.

3. The graph union of all the fragments contains no directed cycles. Nodes are considered identical in forming the graph union if the associated hypothesis types are the same.

Figure 2 shows the graph union of the MEBN represented in Figure 1. The units of replication for this model are units *u* representing possible platoon-sized groups, reports *r* representing observable information about these platoon-sized groups, and parent units *p* representing company-sized groups made up of platoon-sized groups. This is a highly simplified model representing only a few aspects of the domain, but it is sufficient to illustrate the basic concept of an MEBN. The model of Figure 2 can be replicated for arbitrarily many company sized units, constituent platoons, and reports. Thus, this set of fragments constitutes a coherent probability model over an arbitrarily complex web of interrelated hypotheses. Sample spaces of hundreds to thousands of random variables can easily be built up in this manner from even very small knowledge bases.

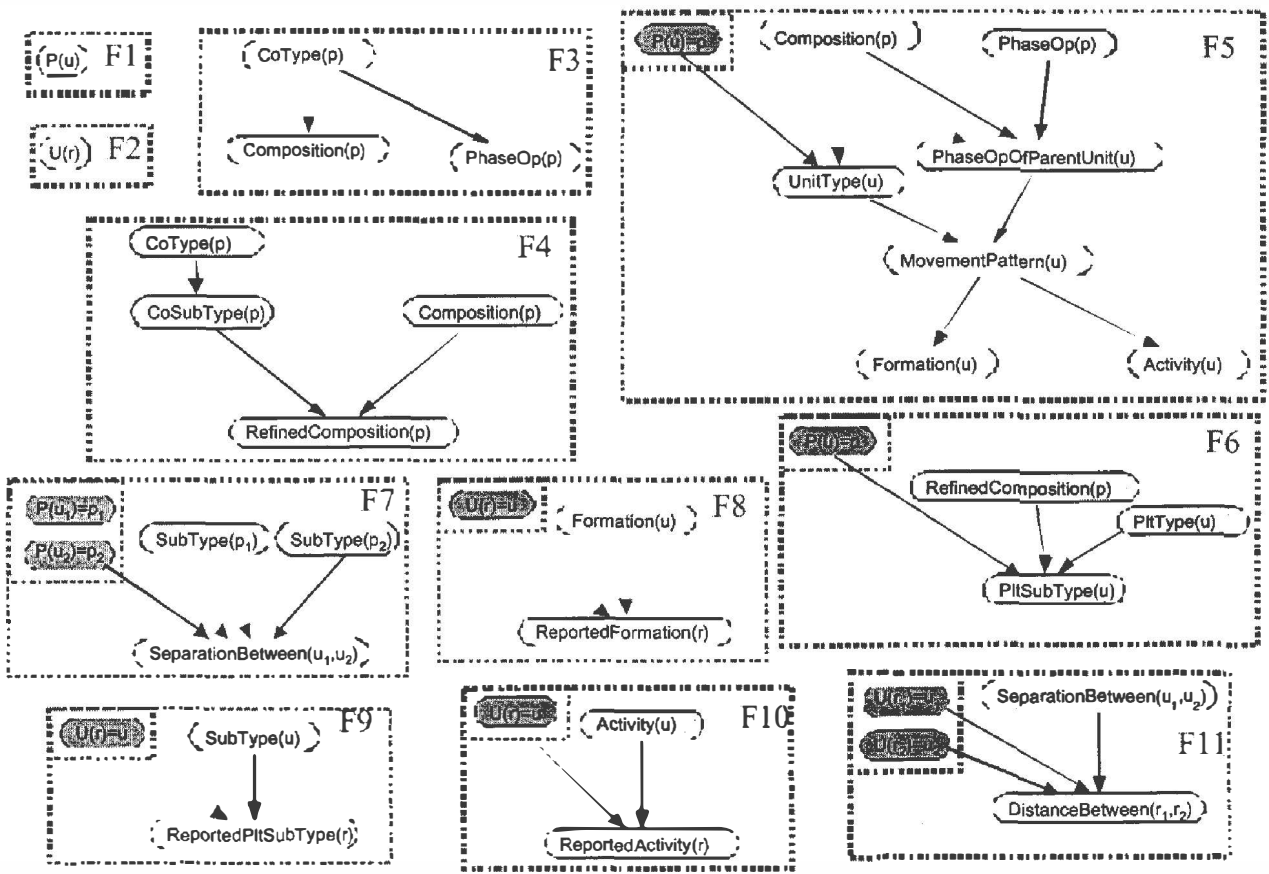

**Figure 1: Multi-Entity Bayesian Network Model**

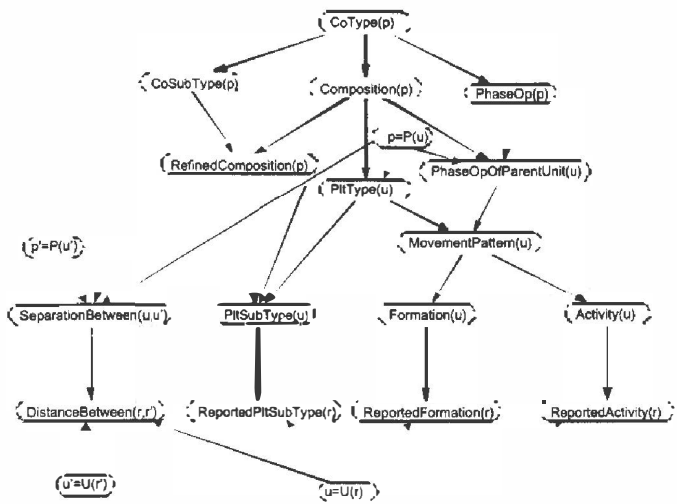

**Figure 2: Graph Union of MEBN of Figure 1**

## 4. SITUATION-SPECIFIC MEBNs

The canonical problem we consider is to infer the values of a set of target hypotheses given a set of evidence hypotheses. For example, we might be tasked with identifying the unit type and activity of all military units in a given geographic region. More generally, let $E(x,y)$ be a set of evidence hypothesis instances whose values are given. Let $T(x,Z)$ denote a set of target hypotheses whose values we wish to infer. The label $x$ denotes the actual hypothesis instances referred to by the identifying attributes shared by target and evidence hypotheses. The label $Z$ refers to the unknown values of identifying attributes the target hypotheses do not share with the evidence variables. Our objective is to compute the response to a query of the form

$$Q: [ \ P(T(x,Z)|E(x,y)=e) \ = \ ? \ ]. \tag{1}$$

To consider an example, suppose in our situation assessment problem we have two reports $r_1$ and $r_2$ which provide type information about the entity to which they refer. That is, our evidence consists of the information

RepPltSubType$(r_1) = t_1$ and
RepPltSubType$(r_2) = t_2$.

We might be interested in the actual type of the units giving rise to these reports and of their parent units. That is, we might be interested in PltSubType($U$) and CoSubType($P$), where $U$ and $P$ are the unknown platoon sized unit(s) and parent company sized unit(s) of interest.

This question can be answered by constructing a situation-specific network (Mahoney and Laskey, 1998) from the MEBN of Figure 1. A situation-specific network is a minimal query-complete network: a (perhaps partially specified) BN that contains sufficient information to compute the response to the query. A situation-specific network can be created by beginning with any query-complete network and pruning barren nodes, nuisance nodes, and nodes $d$-separated by the evidence variables from the target variables. Alternatively, it can be built up incrementally from a knowledge base of BN fragments. Most incremental model construction algorithms apply a variant of what Mahoney and Laskey (1998) call simple bottom-up construction. Simple bottom-up construction builds a BN by constructing upward from the evidence and target variables, adding parents to each node until either a root node or an evidence node is reached. The knowledge base of BN fragments provides the distributions and links to the parents for all the variables. If marginal distributions at nodes are cached, then the simple bottom-up construction algorithm can be modified to terminate when a nuisance node is reached (Lin and Druzdzel, 1997; Mahoney and Laskey, 1998).

Consider the problem of constructing a situation-specific network for the first report in our example. We receive report $r_1$, hypothesize a platoon sized unit $u_1$ that generated it, and hypothesize a company sized unit $p_1$ to which it belongs. Applying network construction results in the situation-specific network of Figure 3a. No association hypotheses need be explicitly represented. Only fragments F3, F4, F5, F6 and F9 were needed to construct this network. Nodes in the retrieved fragments relating to movement pattern and phase of operation were removed as irrelevant to the query.

Association hypotheses become relevant when the second report is received. This report might refer to the already-referenced unit $u_1$ or it might refer to a different unit $u_2$. If it refers to a different unit, it might be a sub-unit of the same parent unit $p_1$ or it might have a different parent $p_2$. Figure 3b and 3c show the constructed networks for the two association hypotheses we could make for $r_2$. Notice that these two hypotheses share a fairly large sub-network. Depending on the amount of shared structure, it may be more computationally efficient to combine the hypothesized networks into a single network as shown in Figure 3d. In this network, the parents of any child of an association hypothesis include multiple replicates of the parents referring to the hypothesized attribute, one for each hypothesized association. Explicit representation of the CPT for such a node would rapidly become unwieldy, but hypothesis-specific independence can be exploited for both computational and representational efficiency. Constructing the graph of Figure 3d is a straightforward application of the network construction algorithm described in Mahoney and Laskey (1998), with minor modifications to handle association hypotheses. These modifications are summarized as follows:

*Hypothesis enumeration and data association:* Retrieval of a fragment containing association hypotheses triggers enumeration of candidate values for the association hypothesis. The candidates include all existing hypothesis instances which match the type of the association hypothesis. In addition, a new hypothesis instance may be created specifically to account for this association hypothesis. Continuing with our situation assessment example, if we were to receive a third report of a platoon-sized group, we would consider associating it with $u_1$, $u_2$, or a previously unreported new unit $u_3$.

Typically, explicitly including all possibilities for each association hypothesis would be intractable. However, most of these hypotheses are far too improbable to merit explicit consideration. Heuristic filters can be applied to narrow down the hypotheses to a manageable number. This process is discussed in more detail in the next section.

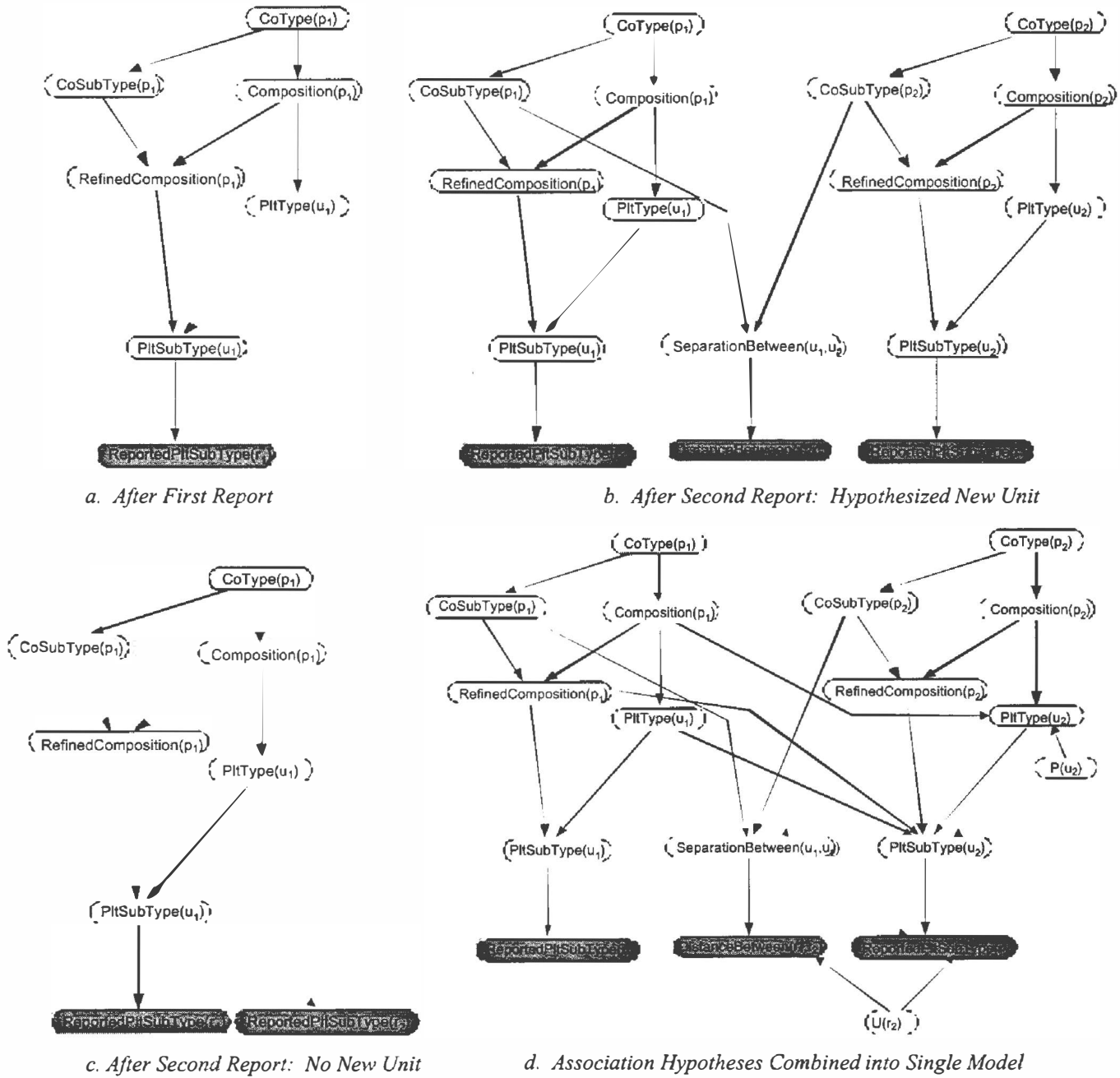

a. After First Report

b. After Second Report: Hypothesized New Unit

c. After Second Report: No New Unit

d. Association Hypotheses Combined into Single Model

**Figure 3: Situation-Specific Network with Association Uncertainty**

*Network construction.* Once a set of candidate hypotheses has been enumerated, a copy of the network fragment is made for each candidate and the fragments are combined. We need to introduce special machinery for reasoning about hypothesis instances generated by incorrect association hypotheses. For example, how should the system treat statements about the type, activity, formation, etc. of unit $u_2$ under the hypothesis that report $r_2$ is associated with unit $u_1$? This issue arises in any system that reasons about unknown numbers of entities in unknown relationship with each other. We introduce a special symbol *, meaning "not relevant to query," to refer to incorrect association hypotheses and consequences derived from them. The value * is treated as follows:

- Any node that is a descendent of an association hypothesis must have * as a possible value.

- Any co-parent of an association hypothesis must have * as a possible value.

- At least one ancestor of any co-parent of an association hypothesis must have * as a possible value.

- Distributions for * may be assigned explicitly by the knowledge engineer or handled automatically. Automatic handling follows these rules:

  - For non-root nodes, the CPT stores distributions for all values except *.

  - A root node with possible value * has a type-specific probability assigned to *. This value can be tuned by the system designer. It represents the *a priori* probability that any nominated association hypothesis of this type will turn out to be query-relevant.

  - If any parent of a node has value *, the node has value * with probability 1. Distributions conditioned on non-* values of the parent nodes are assigned according to the fragment CPT.

A final consideration is the assignment of probability distributions to nodes with association hypotheses as parents. Let $A(x,y)$ be the parent of a node $B(x,z)$ with association hypothesis $Y(z_j)$. The node $B(x,z_j)$ in the constructed network will have as parent the association hypothesis node $Y(z_j)$ and a copy of $A(x,y_k)$ for each state $y_k$ of $Y(z_j)$. The probability distribution for $B(x,z_j)$ simply picks out the fragment local distribution for the copy of $A(x,y_k)$ associated with the state $y_k$ of $Y$.

In addition to representing associations of a report with the wrong unit or platoons with the wrong company, the value * can also be used to represent false alarms, or reports not generated by any unit of interest. Another function of * is to represent errors in subtyping. For example, to reason about whether a platoon is an armor or mechanized platoon, a hypothesis instance is created for each subtype. All attributes represented at the subtype level have value * in the armored platoon hypothesis instance when the mechanized platoon hypothesis is assumed, and vice versa. The value * can also be used to reason about number uncertainty. For example, we might create three platoon instances for a company, but assign all attributes of the third platoon values of * under the hypothesis that the company has suffered casualties and has only two platoons. Indeed, associations to the wrong unit, false alarms, subtyping errors and number errors are all forms of mis-association, and thus it is natural to treat them in a unified way.

## 5   MANAGING HYPOTHESES IN MEBNS

If each company contains 3 maneuver platoons and a company headquarters platoon, and each platoon contains about four vehicles, then a five company scenario involves approximately 80 vehicles. Assuming that false alarms are slightly more probable than unobserved vehicles, the scenario would require processing approximately 90 vehicle reports. Without considering missed detections or false alarms, there are over 2.5 million ways to form a 4-vehicle platoon hypothesis from 90 vehicle reports, over 2.1 million ways to form a second 4-vehicle platoon hypothesis from the remaining 86 reports, and over 1.7 million ways to form a third 4-vehicle platoon hypothesis. Clearly, brute force enumeration of all association hypotheses is an infeasible strategy. Fortunately, the vast majority of the possible association hypotheses are wildly improbable. Because vehicles in the same platoon tend to stay near each other, a distance threshold can be applied to screen out alternatives too far apart to be part of the same platoon. For example, there are only 210 4-vehicle platoon hypotheses for a cluster of 10 vehicles meeting the distance threshold. Thus, distance thresholding can cut down the combinatorics by many orders of magnitude. Nevertheless, several hundred association hypotheses per report set is still far too many to represent explicitly. However, our objective is not to draw accurate inferences about which vehicle reports are associated which platoons and companies, but rather to infer certain summary features. Details of how hypotheses are associated to reports is unimportant if the number, approximate locations, types, and activities of companies can be inferred. Thus, the problem of searching over and evaluating an intractably large number of similar hypotheses can be replaced by the problem of enumerating a few candidate hypotheses that adequately represent the population of hypotheses for the purpose of estimating the features important to consumers of the model.

In our architecture, network construction is preceded by a pre-processing step that nominates promising candidate hypotheses. Clusters of reports trigger firing of *suggestors*, which are rules mapping situation features to situation hypotheses. The features used by suggestors are input reports and details of the current situation model. Suggestors are obtained from the MEBN model by identifying combinations of observable features that are highly diagnostic of hypotheses of interest and selecting threshold values that weed out the most improbable hypotheses while retaining hypotheses worthy of further evaluation. Suggestors may perform arbitrary computations, such as clustering or invoking a decision model.

When a suggestor fires, it initiates a process to identify hypotheses to be explicitly evaluated by the constructed network. Hypotheses may be generated for previously unobserved entities, or queries may be performed against the current hypothesis set to find possible associations between incoming reports / hypotheses and existing hypotheses. Associations can be definite or hypothetical. We currently have implemented only a single hypothesis approach for hypothetical associations. That is, only two values are considered for each association hypothesis: the possibility nominated by the suggestor or the value *, representing a false alarm. We also use * to represent subtyping uncertainty. We plan to consider multiple association hypotheses in future work. Hypothesis enumeration is followed by construction of the assessment model and evaluation of candidate hypotheses to determine the level of evidential support.

The refinement decision cycles back through the process. For example, a platoon-sized group hypothesis may be refined to a mechanized infantry platoon by the assessment model. This refinement could result in triggering a different suggestor. Posterior probabilities on * are useful for control of the interleaved cycle of construction and evaluation. If a hypothesis instance has high probability of *, it may be a candidate for pruning or combining with similar hypotheses. The network construction algorithm may also have heuristic steps in which construction terminates at non-root nodes with a default distribution unless evaluation produces a high enough probability on non-* values to justify further construction. The interpretation of * as "not relevant to query" suggests treating nodes with high belief in * in a similar manner to nuisance nodes. Upward construction might terminate at such nodes and belief in * is assigned by default to the marginal distribution for non-* states.

The architecture in which suggestors trigger construction of assessment models is designed to achieve a high probability of detection while tolerating a moderate false alarm rate. The assessment model is applied after detection to refine the inference and reduce the false alarm rate to acceptable levels. Detection thresholds can be adjusted at design time to balance processing load against performance. The engine that manages and carries out the model construction and inference process is domain independent. The domain dependent elements are the BN fragments and the library of suggestors.

# 6. DISCUSSION

Multi-entity Bayesian networks represent knowledge in domains in which the number and relationships among domain entities cannot be pre-specified. Query processing in an MEBN requires knowledge-based model construction. In the presence of uncertainty about the association of hypotheses to evidence about the hypotheses, model construction is typically intractable. We described an architecture for model construction from MEBN knowledge bases. Suggestors are used to nominate hypotheses for consideration. Knowledge based model construction is used to construct an assessment model that refines the situation estimate. This architecture follows a strategy similar to Goldman (1990) and to the strategies followed by typical multitarget tracking systems (e.g., Stone, et al., 1999). Our suggestors correspond to the marker passer used by Goldman to nominate candidate explanations, and to the gating thresholds and data association methods applied in multitarget tracking systems. Our assessment models correspond to the Bayesian network used by Goldman to evaluate explanations nominated by the marker passer, and to the state association and temporal projection modules of multi-target tracking systems. The common feature in all these architectures is a pre-processing step that trades off a moderate false alarm rate for fast computation and a low miss rate, followed by a more computationally demanding main process to weed out false alarms and compute accurate responses to queries. This general approach depends on spatio-temporal locality of the features used by the pre-processor. The locality assumption is required in order to achieve a low miss rate in pre-processing while keeping the false alarm rate within bounds the main process can handle.

## Acknowledgments

Research for this paper was partially supported by DARPA & AFRL contract F33615-98-C-1314, Alphatech subcontract 98036-7488. We thank Bruce D'Ambrosio, Masami Takikawa, and Dan Upper for software design and implementation, Gary Patton of Veridian for domain expertise, Tod Levitt of IET for useful discussion, and Otto Kessler of DARPA for technical vision.